\documentclass[11pt,a4paper]{article}
\usepackage[hyperref]{acl2021}
\usepackage{times}
\usepackage{latexsym}
\usepackage{graphicx}

\usepackage{acronym}
\usepackage{booktabs}

\expandafter\def\expandafter\UrlBreaks\expandafter{\UrlBreaks
  \do\a\do\b\do\c\do\d\do\e\do\f\do\g\do\h\do\i\do\j%
  \do\k\do\l\do\m\do\n\do\o\do\p\do\q\do\r\do\s\do\t%
  \do\u\do\v\do\w\do\x\do\y\do\z\do\A\do\B\do\C\do\D%
  \do\E\do\F\do\G\do\H\do\I\do\J\do\K\do\L\do\M\do\N%
  \do\O\do\P\do\Q\do\R\do\S\do\T\do\U\do\V\do\W\do\X%
  \do\Y\do\Z}

\newacro{LIP}{Language Invariant Property}
\newacro{LIPs}{Language Invariant Properties}  

\usepackage{microtype}
\usepackage[compact]{titlesec}

\aclfinalcopy 

\title{Language Invariant Properties in Natural Language Processing}

\author{Federico Bianchi, Debora Nozza, Dirk Hovy \\
  Bocconi University \\
  Via Sarfatti 25 \\
  Milan, Italy \\
  \texttt{\{f.bianchi,debora.nozza,dirk.hovy\}@unibocconi.it} }
\date{}

\begin{document}
\maketitle
\begin{abstract}
Meaning is context-dependent, but many properties of language (should) remain the same even if we transform the context. For example, sentiment, entailment, or speaker properties should be the same in a translation and original of a text.
We introduce \textbf{language invariant properties}: i.e., properties that should not change when we transform text, and how they can be used to quantitatively evaluate the robustness of transformation algorithms.
We use translation and paraphrasing as transformation examples, but our findings apply more broadly to any transformation.
Our results indicate that many NLP transformations change properties like author characteristics, i.e., make them sound more male.
We believe that studying these properties will allow NLP to address both social factors and pragmatic aspects of language. We also release an application suite that can be used to evaluate the invariance of transformation applications.
\end{abstract}


\section{Introduction}
The meaning of a sentence is influenced by a host of factors, among them who says it and when: ``That was a sick performance'' changes meaning depending on whether a 16-year-old says it at a concert or a 76-year-old after the opera.\footnote{Example due to Veronica Lynn.}
However, there are several properties of language that do (or should) not change when we \textit{transform} a text (i.e., change the surface form of it to another text, see also Section \ref{sec:lips}). If the text was written by a 25-year-old female it should not be perceived as written by an old man after we apply a paraphrasing algorithm. 
The same goes for other properties, like sentiment: A positive message like ``good morning!'', posted on a social media, should be perceived as a positive message, even when it is translated into another language.\footnote{See for example past issues like \url{https://www.theguardian.com/technology/2017/oct/24/facebook-palestine-israel-translates-good-morning-attack-them-arrest}}
We refer to these properties that are unaffected by transformations as \textbf{\ac{LIPs}}.

\ac{LIPs} preserve the semantics and pragmatic components of language. I.e., these properties are not affected by transformations applied to the text.
For example, we do not expect a summary to change the topic of a sentence. 

\textbf{Paraphrasing}, \textbf{summarization}, \textbf{style transfer}, and \textbf{machine translation} are all NLP transformation tasks that we expect to respect the \ac{LIPs}. If they do not, it is a strong indication that the system is picking up on spurious signals, and needs to be recalibrated. For example, machine translation should not change speaker demographics or sentiment, paraphrasing should not change entailment or topic.

\begin{figure*}
\centering
\includegraphics[width=0.8\textwidth]{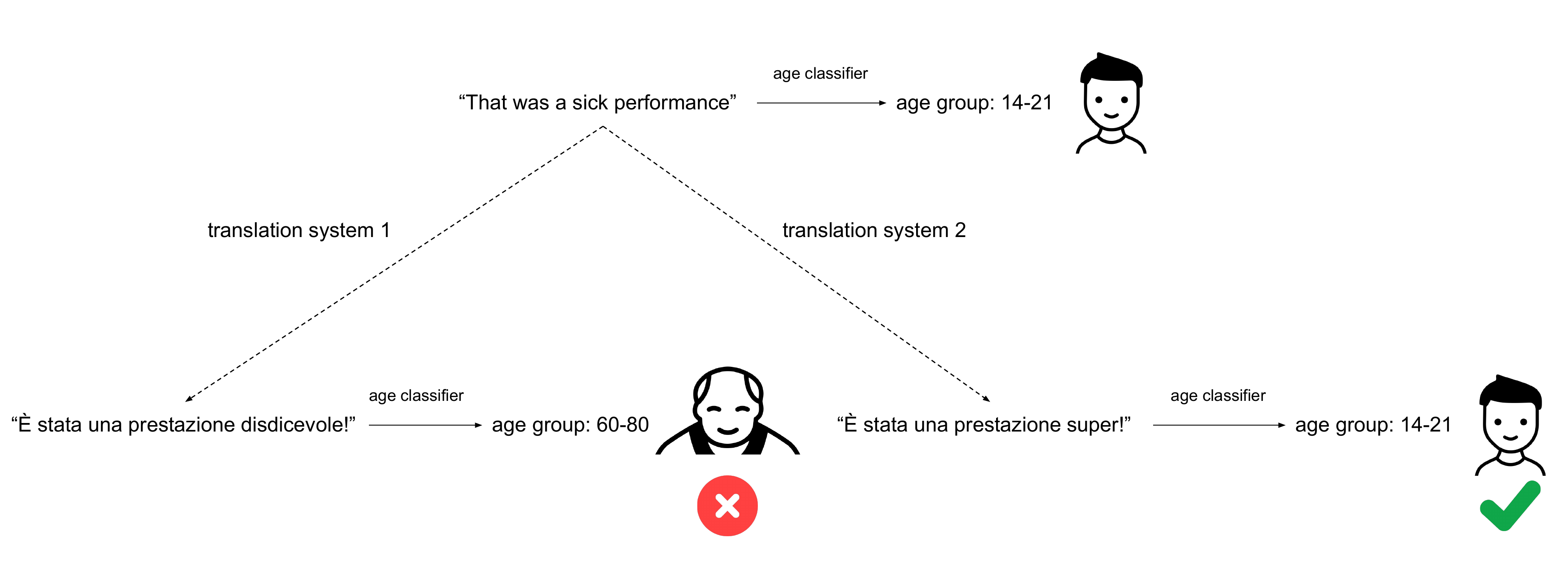}
\caption{Author age is a \ac{LIP}. Translation system 1 fails to account for this and provides a translation that can give the wrong interpretation to the sentence. Translation system 2 is instead providing a more correct interpretation.}
\label{fig:LIP:example}
\end{figure*}

However, an important question is, what happens if a transformation \textit{does} violate invariants?
Violating invariants is similar to breaking the cooperative principle ~\cite{grice1975logic}: if we do it deliberately, we might want to achieve an effect. For example, \newcite{reddy-knight-2016-obfuscating} showed how words can be replaced to obfuscate author gender and thereby protect their identity.
Style transfer can therefore be construed as a deliberate violation of \ac{LIPs}.
In most cases, though, violating a \ac{LIP} will result in an unintended outcome or interpretation of the transformed text: for example, violating \ac{LIPs} on sentiment will generate misunderstanding in the interpretation of messages. Any such violation might be a signal that models are not ready for production~\cite{bianchi-hovy-2021-gap}.

We know that cultural differences can make it more difficult to preserve \ac{LIPs} \cite{hovy-yang-2021-importance}: it might not be possible to effectively translate a positive message into a language that does not share the same appreciation/valence for the same things. However, this is a more general limitation of machine translation. The speaker's intentions are to keep the message consistent - in term of \ac{LIPs} - even when translated.

We define the concept of \ac{LIPs}, but also integrate insights from \newcite{hovy-etal-2020-sound}, defining an initial benchmark to study \ac{LIPs} in two of the most well-known transformation tasks: machine translation and paraphrasing. We apply those principles more broadly to transformations in NLP as a whole.

\paragraph{Contributions.} We introduce the concept of \ac{LIPs} as those properties of language that should not change during a transformation. We propose an evaluation methodology for LIPs and release a benchmark that can be used to test how well translation systems can preserve LIPs for different languages.

\section{Language Invariant Properties}
\label{sec:lips}
To use the concept of \ac{LIPs}, we first need to make clear what we mean by it. We formally define  \ac{LIPs} and transformations below.

Assume the existence of a set $S$ of all the possible utterable sentences. Let's define $A$ and $B$  as subsets of $S$. These can be in the same or different languages. Now, let's define a mapping function $$t: A \rightarrow B$$
i.e., $t(\cdot)$ is a \textbf{transformation} that changes the surface form of the text $A$ into $B$.

A \textit{language property} $p$ is a function that maps elements of $S$ to a set $P$ of property values. $p$ is \textbf{invariant} if and only if 
$$p(a) = p(t(a)) = p(b)$$
where $a \in A, b \in B$, and $t(a) = b$. I.e., if applying $p(\cdot)$ to both an utterance and its transformation still maps to the same property.
We do not provide an exhaustive list of these properties, but suggest to include at least \textbf{meaning}, \textbf{topic}, \textbf{sentiment}, \textbf{speaker demographics}, and \textbf{logical entailment}. 
\\\\
\ac{LIPs} are thus based on the concept of transformations in text. Machine translation is a salient example of a transformation, and probably the prime example of a task for which \ac{LIPs} are important. MT can be viewed as a transformation between two languages where the main fundamental \ac{LIP} that should not be broken is meaning. 

However, \ac{LIPs} are not restricted to MT, but have broader applicability, e.g., in style transfer. In that case, though, some context has to be defined. When applying a \textit{formal} to \textit{polite} transfer, this function is by definition \textit{not} invariant anymore. Nonetheless, many other properties should not be influenced by this transformation.
Finally, for paraphrasing, we have only one language, but we have the additional constraint that $t(a) \neq a$. For summarization, the constraint instead is that $len(t(a)) < len(a)$. 

\ac{LIPs} are also what make some tasks in language more difficult than others: for example, data augmentation~\cite{feng-etal-2021-survey} cannot be as easily implemented in text data as in image processing, since even subtle changes to a sentence can affect meaning and style. Changing the slant or skew of a photo will still show the same object, but e.g., word replacement easily breaks \ac{LIPs}, since the final meaning of the final sentence and the perceived characteristics can differ. Even replacing a word with one that is similar can affect \ac{LIPs}. For example, consider machine translation with a parallel corpus: ``the dogs are running'' can be paired with the translation ``I cani stanno correndo'' in Italian. If we were to do augmentation, replacing \textit{dogs} with its hyperonym ``animals'' does not corrupt the overall meaning, as the new English sentence still entails all that is entailed by the old one. However, the Italian example is no longer a correct translation of the new sentence, since ``cani'' is not the word for animals.

\ac{LIPs} are also part of the communication between speakers. The information encoded in a sentence uttered by one speaker contains \ac{LIPs} that are important for efficient communication, as misunderstanding a positive comment as a negative one can create issues between communication partners.

Note that we are not interested in evaluating the \textit{quality} of the transformation (e.g., the translation or paraphrase). There are many different metrics and evaluation benchmarks for that \cite[BLEU, ROUGE, etc.:][]{papineni-etal-2002-bleu,lin-2004-rouge}. Our analysis concerns another aspect of communication.

The general ideas behind \ac{LIPs} shares some notions with the \textit{Beyond Accuracy Checklist} by ~\newcite{ribeiro-etal-2020-beyond}. However, \ac{LIPs} 
evaluate how well fundamental properties of discourse are preserved in a transformation, the CheckList is made to guide users in a fine-grained analysis of the model performance to better understand bugs in the applications with the use of templates. As we will show later, \ac{LIPs} can be quickly tested to any new annotated dataset. Some of the checklist' tests, like \textit{Replace neutral words with other neutral words} can be seen as \ac{LIPs}. Nonetheless, we think the two frameworks are complementary.

\section{Evaluating Transformation Invariance}

For ease of reading, we will use translation as an example of a transformation in the following. However, the concept can be applied to any of the transformations we mentioned above. 

We start with a set of original texts $O$ to translate and a translation model from the source language of $O$ to a target language. To test the transformation wrt a \ac{LIP}, $O$ should be annotated with that language property of interest. We also need a classifier for the \ac{LIP}. For example, a \ac{LIP} classifier could be a gender classifier that given an input text returns the inferred gender of the speaker. Here, we need one cross-lingual classifier, or two classifiers, one in the source and one in the target language. For all other transformations, which stay in the same language, we only need one classifier.
(Paraphrasing or summarization can be viewed as a transformation from English to English).

After we translate the test set, we can run the  classifier on the $O$ data, to get its predictions ($PO$). We then run the classifier on the transformed data, generating the predictions on the transformed data ($PT$).

We can then compare the difference between the distribution of the \ac{LIP} in the original data and either prediction. I.e., we compare the differences of $O-PO$ to $O-PT$ to understand the effect of the transformations. 

Note that we are \textit{not} interested in the actual performance of the classifier, but in the difference in performance on the two data sets.  We observe two possible phenomena:

\begin{itemize}
    \item If there is a \textit{classifier bias}, both the predictions based on the original language and the predictions based on the translations should be skewed in the \textbf{same} direction wrt the distribution in $O$. E.g., for gender classification, both classifiers predict a higher rate of male authors in the original and in the translated text. 
    \item  Instead, if there is a \textit{transformation bias}, then the distribution of the translated predictions should be skewed in a different direction than the one based on the original language. E.g., the gender distribution in the original language should be less skewed than the gender ratio in the translation. 
\end{itemize}

As we will see in the next Section, it is possible to use divergence metrics (like the KL divergence) to quantify the difference between the classifiers. To reduce one of the possible sources of bias, the classifier should be trained with data that comes from a similar distribution to the one used at test time, ideally from the same collection.

\section{Benchmark Tool}

We release an extensible benchmark tool\footnote{\url{https://github.com/MilaNLProc/language-invariant-properties}} that can be used to quickly assess a model's capability to handle \ac{LIPs}. 

\subsection{Datasets}

Here, we evaluate machine translation and paraphrasing as tasks. Our first release of this benchmark tool contains the datasets from~\newcite{hovy-etal-2020-sound}, annotated with gender\footnote{The dataset comes with binary gender, but this is not an indication of our views or the capabilities of the benchmark tool.} and age categories, and the SemEval dataset from~\newcite{mohammad-etal-2018-semeval} annotated with emotion recognition. Moreover, we include the English dataset from HatEval~\cite{basile-etal-2019-semeval} containing tweets for Hate Speech detection. These datasets come with training splits that we use to train the classifiers.

Nonetheless, the benchmark we provide can be easily extended with new datasets encoding other \ac{LIPs}.

\begin{figure}
\centering
\includegraphics[width=1\columnwidth]{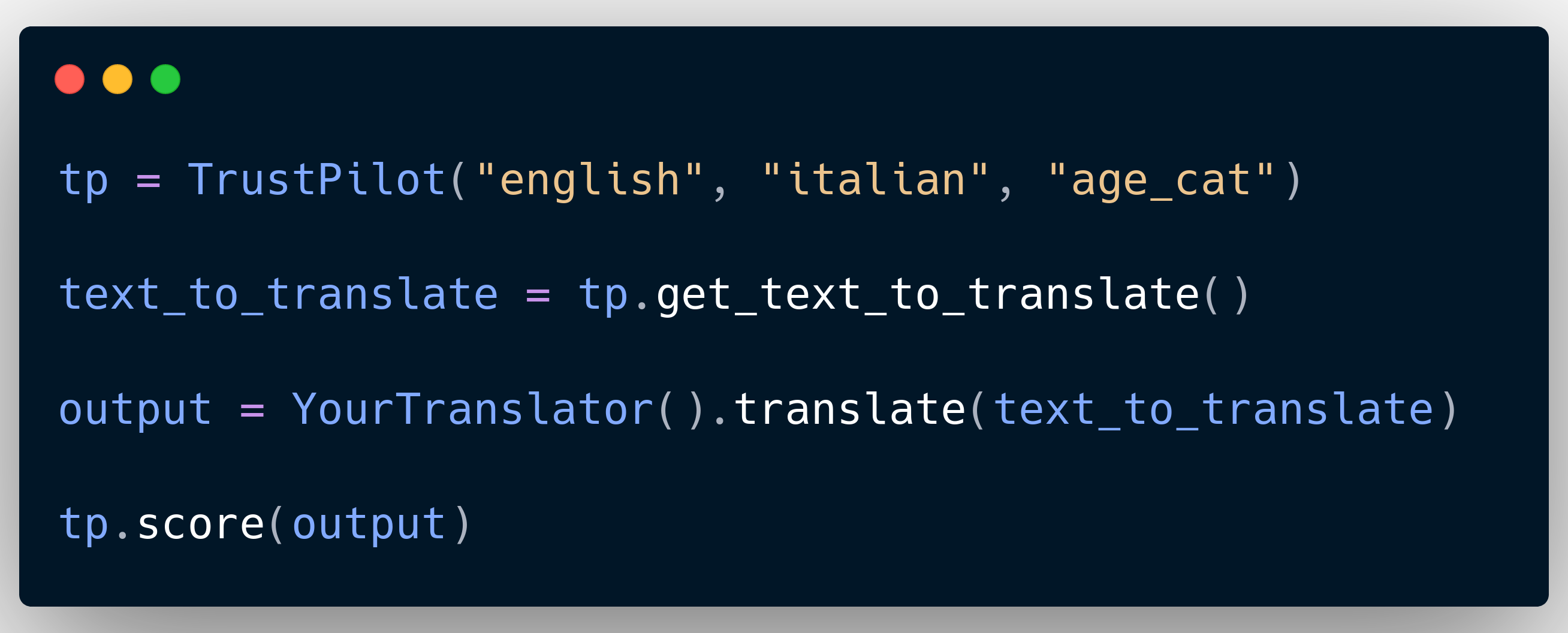}
\caption{The benchmark has been designed to provide a high-level API that can be integrated in any transformation pipeline. Users can access the dataset text, transform, and score it.}
\label{fig:carbon}
\end{figure}

\paragraph{TrustPilot}

The dataset is a subset of the one originally presented by ~\newcite{hoyv2015user} and contains TrustPilot reviews in English, Italian, German, French, and Dutch with demographic information about the user age and gender. Training data for the different languages consists of 5,000 samples (2,500 for the male and 2,500 for the female genders). More information can be found in the original work~\newcite{hovy-etal-2020-sound}. The dataset can be used to evaluate the \ac{LIPs} \textsc{author-gender} and \textsc{author-age}.

\paragraph{HatEval}
We use the data from HatEval~\cite{basile-etal-2019-semeval} considering only English tweets. We take the training (9,000 examples) and test set (3,000 examples). Each tweet comes with a value that indicates if the tweet contains hate speech. The dataset can be used to evaluate the \ac{LIP} \textsc{hatefulness}.

\paragraph{Affects in Tweets}
We also use the Affect in Tweets dataset (AiT)~\cite{mohammad-etal-2018-semeval}, which contains tweets annotated with emotions. We will use this dataset to test how sentiment and emotions are affected by translation. We do not expect much variation in the distribution, since both emotion and sentiment should be efficiently translated; this is exactly what we would like our evaluation framework to show.

To reduce the number of possible classes to predict we only kept emotions in the set \{\textit{fear, joy, anger, sadness}\} to allow for future comparisons with other datasets. This subset is also easy to map to other labels to create a sentiment analysis task: we map \textit{joy} to \textit{positive} and the other emotions to \textit{negative} following \newcite{bianchi-etal-2021-feel}. The data we collected comes in English (train: 4,257, test: 2,149) and Spanish (train: 2,366, test: 1,908). The dataset can be used to evaluate the \ac{LIP} \textsc{sentiment}.

\subsection{Classifiers}

As default classifier we use a Logistic Regression models with L2 regularization over 2-6 TF-IDF character-grams like~\newcite{hovy-etal-2020-sound}. We also provide the use of embedding models from SBERT~\cite{reimers-gurevych-2019-sentence} to generate representations of documents that we can then classify with logistic regression (see Appendix). The two classification methods are referred to as TF (TF-IDF) and SE (Sentence Embeddings). The framework supports the use of any custom classifiers.

\subsection{Scoring}

Standard metrics for classification evaluation can be used to assess how much \ac{LIPs} are preserved during a transformation. Following~\newcite{hovy-etal-2020-sound} we use the KL divergence to compute the distance - in terms of the distribution divergence - between the two predicted distributions. The benchmark also outputs the $X^2$ test to assess if there is a significant difference in the predicted distributions. It is also possible to look at the plots of the distribution to understand the effects of the transformations (see following examples in Figures~\ref{fig:ait},~\ref{fig:para:trust} and~\ref{fig:hatespeech}).

\section{Evaluation}

We evaluate four tasks, i.e., combinations of transformations (translation and paraphrasing) and \ac{LIPs} (gender, sentiment, and hatefulness). The combination is determined by the availability of the particular property in the respective dataset.

\subsection{TrustPilot Translation - LIP: \textsc{author-gender}}

We use the TrustPilot dataset to study the author-gender \ac{LIP} during translation. We use the google translated documents provided by the authors. We are essentially recomputing the results that appear in the work by~\newcite{hovy-etal-2020-sound}. As shown in Table~\ref{tab:results:trustpilot}, our experiments confirm the one in the paper: it is easy to see that the translations from both Italian and German are more likely to be predicted as male. 

\begin{table*}[]
    \centering
    \begin{tabular}{c|ccccccc} \toprule
        Method & L1 &  L2 & $KL_{O,PO}$ & $KL_{O,PT}$ & Dist O & Dist PO & Dist PT \\ \midrule
        SE & IT & EN & 0.004 & 0.034 & M: 0.52, F: 0.48 & M: 0.56, F: 0.44 & M: \textbf{0.64}, F: 0.36 \\ 
        TF & IT & EN & 0.000 & 0.034 & M: 0.52, F: 0.48 & M: 0.53, F: 0.47 & M: \textbf{0.64}, F: 0.36 \\ \midrule
        SE & DE & EN & 0.000  & 0.030 & M: 0.50, F: 0.50 & M: 0.49, F: 0.51 &  M: \textbf{0.61}, F: 0.39  \\ 
        TF & DE & EN & 0.001 & 0.022 & M: 0.50, F: 0.50 & M: 0.52, F: 0.48 &  M: \textbf{0.60}, F: 0.40\\ \bottomrule
     \end{tabular}
    \caption{Results on the TrustPilot dataset when translating from Italian (IT) and German (DE) to English. We use a logistic regression classifier with TF-IDF (TF) and a (cross-lingual) embedding model with a logistic regression classifier (SE). I.e., translation breaks the \ac{LIP} \textsc{author-gender}}
    \label{tab:results:trustpilot}
\end{table*}

\subsection{AiT Translation - LIP:  \textsc{Sentiment}}

\begin{figure}
\centering
\includegraphics[width=1\columnwidth]{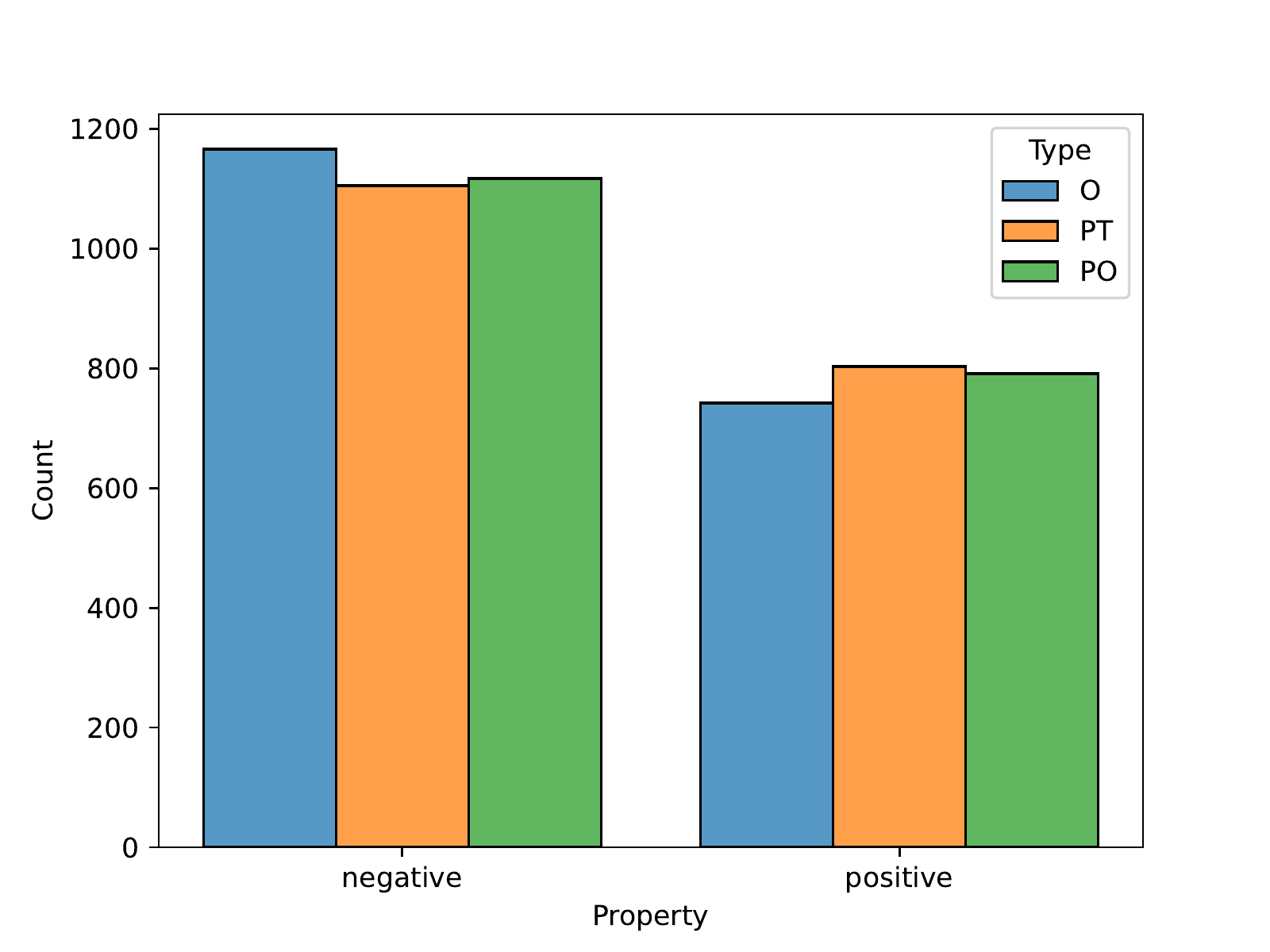}
\caption{Translation on AiT from Spanish to English sentiment analysis. There is little variation of the distribution after the translation. I.e., translation respects the \ac{LIP} \textsc{sentiment}}
\label{fig:ait}
\end{figure}

We use the AiT dataset to test the sentiment \ac{LIP} during translation. We translate the tweets from Spanish to English using DeepL. As shown in Figure~\ref{fig:ait}, sentiment is a property that seems to be easily kept during translations. This is expected, as sentiment is a fundamental part of the meaning of a sentence and has to be translated accordingly.

\subsection{TrustPilot Paraphrasing - LIP: \textsc{author-gender}}

\begin{figure}
\centering
\includegraphics[width=1\columnwidth]{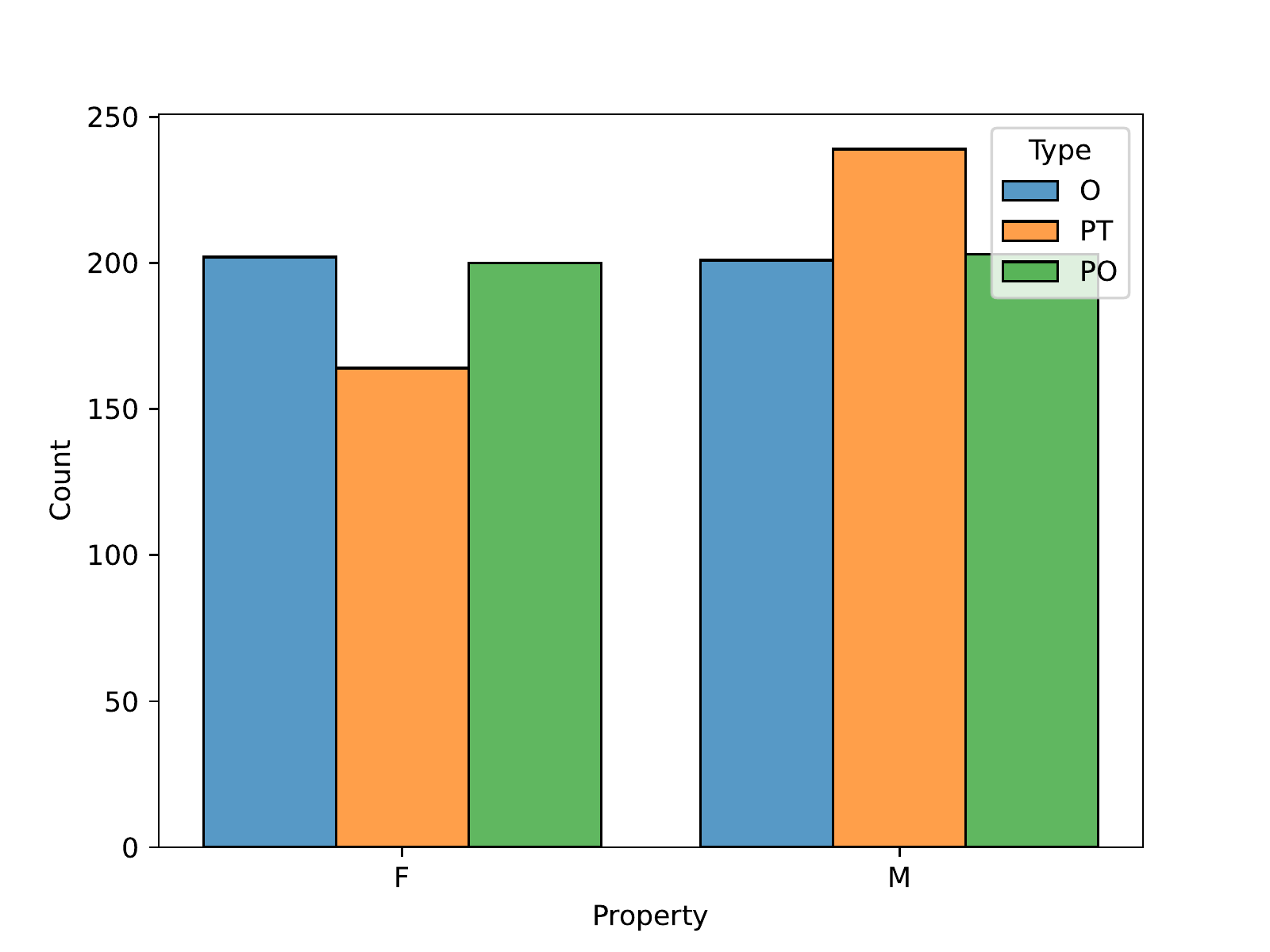}
\caption{Paraphrasing on TrustPilot English data. The number of male-sounding messages increases after paraphrasing. I.e., paraphrasing breaks the \ac{LIP} \textsc{author-gender}}
\label{fig:para:trust}
\end{figure}

We use the TrustPilot dataset to test the author-gender \ac{LIP} in the context of paraphrasing. When we apply paraphrasing on the  data, the classifier on the transformed data predicts more samples as male, as shown by the Figure~\ref{fig:para:trust} that plots the distribution ($KL_{O,PT}=0.018$, difference significant for $X^2$ with $p < 0.01$).

\subsection{HatEval Paraphrasing - LIP: \textsc{Hatefulness}}

\begin{figure}
\centering
\includegraphics[width=1\columnwidth]{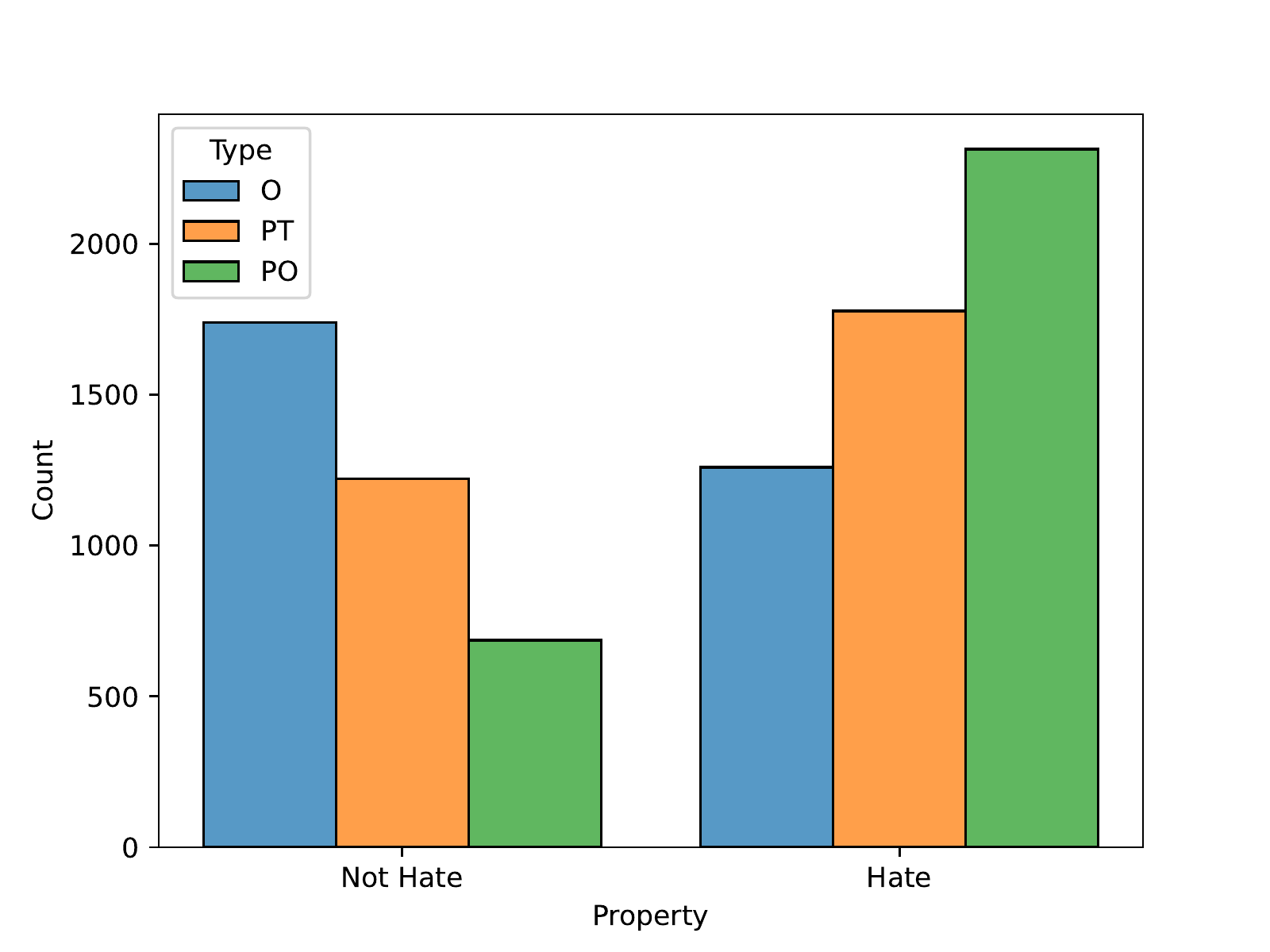}
\caption{Paraphrasing on HatEval English data. The number of hateful tweets is reduced after the paraphrasing. I.e., paraphrasing breaks the \ac{LIP} \textsc{hatefulness}}
\label{fig:hatespeech}
\end{figure}

We use the HatEval data to study the hatefulness \ac{LIP} after paraphrasing. Figure~\ref{fig:hatespeech} shows the change in the predicted distribution: while the classifier predicted a high amount of hateful tweets in $PO$ (a problem due to the differences between the training and the test in HatEval~\cite{basile-etal-2019-semeval,nozza-2021-exposing}), this number is drastically reduced in $PT$, demonstrating that paraphrasing reduce hatefulness.

As an example of paraphrased text, \textit{Savage Indians living up to their reputation} has been transformed to \textit{Indians are living up to their reputation}. While the message stills internalize some hatefulness, the removal of the term \textit{Savage} has reduced its strength. It is important to remark that we are not currently evaluating the quality of the transformation, as this is another task: the results we obtain are in part due to the paraphrasing tool we used,\footnote{\url{https://huggingface.co/tuner007/pegasus_paraphrase}} but they still indicate a limit in the model capabilities.

\section{Limitations}

While \ac{LIPs} make an interesting theoretical concept, they might not always be so easy to generalize to some tasks: For example, translating from Spanish to Japanese has to take account of the cultural differences of the two countries and it might not be possible to reduce these to a set of LIPs. 

The tool we implemented comes with some limitations. We cannot completely removed the learned bias in the classifiers and we always assume that when there are two classifiers, these two perform reliably well on both languages so that we can compare the output. The same goes for the paraphrasing: we assume that the \ac{LIP} classifier perform equally well on both the original and the transformed text.

\section{Conclusion}
This paper introduces the concept of \acl{LIPs}, properties in language that should not change during transformations. We believe the study of \ac{LIPs} can improve the performance of different NLP tasks, like machine translation and paraphrasing. To provide better support in this direction we release a benchmark that can help researchers and practitioners understand how well their models handle \ac{LIPs}.

\section*{Acknowledgments}
We thank Giovanni Cassani and Amanda Curry for the comments on an early draft of this work.

\bibliographystyle{acl_natbib}
\bibliography{acl2021}

\appendix

\section{Models in the Experiments}

\subsection{TrustPilot Paraphrase}

We use the same classifier for the original and the transformed text. We generate the representations with SBERT. The model used is \emph{paraphrase-distilroberta-base-v2}.\footnote{\url{https://sbert.net}}

As paraphrase model, we use a fine-tuned Pegasus~\cite{Zhang2020PEGASUSPW} model, pegasus paraphrase,\footnote{\url{https://huggingface.co/tuner007/pegasus_paraphrase}} that at the time of writing is one of the most downloaded on the HuggingFace Hub.

\subsection{AiT Translation}

We translated the tweets using the DeepL APIs.\footnote{\url{https://deepl.com/}} As classifiers we use the cross-lingual model for both languages, each language has its language-specific classifier. The cross-lingual sentence embedding method used is \emph{paraphrase-multilingual-mpnet-base-v2}, from the SBERT package.

\subsection{TrustPilot Translation}

As translation we use the already translated sentences from the TrustPilot dataset provided by \newcite{hovy-etal-2020-sound}. We use both the TF-IDF based and the cross-lingual classifier, as shown in Table~\ref{tab:results:trustpilot}, each language has its own language-specific classifier. The cross-lingual sentence embedding method used is \emph{paraphrase-multilingual-mpnet-base-v2}, from the SBERT package.

\subsection{HatEval Paraphrasing}

We use the same classifier for the original and the transformed text. We generate the representations with SBERT. Users are replaced with \textit{@user}, hashtags are removed. The model used is \emph{paraphrase-distilroberta-base-v2}.

As paraphrase model, we use a fine-tuned Pegasus~\cite{Zhang2020PEGASUSPW} model, pegasus paraphrase, that at the time of writing is one of the most downloaded on the HuggingFace Hub.

\end{document}